\documentclass[12pt]{article}

\usepackage[english]{babel}
\usepackage{tikz}
\usepackage[letterpaper,top=2cm,bottom=2cm,left=3cm,right=3cm,marginparwidth=1.9cm]{geometry}
\usepackage{pgfgantt}
\usepackage{adjustbox}
\usepackage{afterpage}

\usepackage{mathtools}
\usepackage{setspace} 
\usepackage{amsmath}
\usepackage{graphicx}
\usepackage[colorlinks=true, allcolors=blue]{hyperref}
\usepackage{multirow}
\usepackage{tabularx}
\usepackage{titlesec}
\usepackage{subfig}
\titleformat{\paragraph}
{\normalfont\normalsize\bfseries}{\theparagraph}{1em}{}
\titlespacing*{\paragraph}
{0pt}{3.25ex plus 1ex minus .2ex}{1.5ex plus .2ex}

\begin{document}
\setstretch{1.5} 
\pagenumbering{roman}
\begin{titlepage}
   \begin{center}
       \vspace*{1cm}

       \textbf{\Large Sim2Real Reinforcement Learning for Soccer skills}

       \vspace{0.5cm}
            
    \vspace{0.5cm}
        By
        \\
        \vspace{0.3cm}
        Jonathan Spraggett 
       \\
        jonathan.spraggett@mail.utoronto.ca
        \\
        \vspace{0.8cm}
        Supervisor: Michael Guerzhoy
        \\
        April 14, 2023
        \\
         \vspace{7cm}

        \includegraphics[width=165mm]{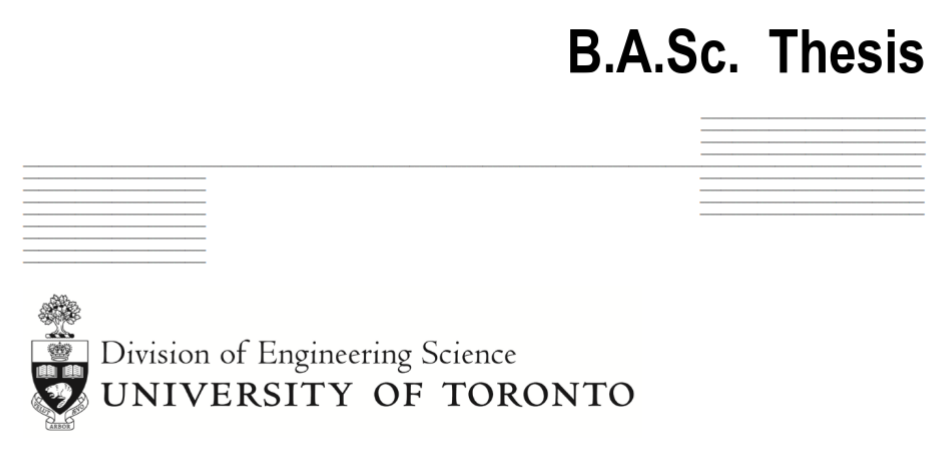}

   \end{center}
\end{titlepage}
\begin{titlepage}
   \begin{center}
       
       This page is intentionally left blank
            
   \end{center}
\end{titlepage}

\begin{titlepage}
   \begin{center}
       \vspace*{1cm}

       \textbf{\Large Sim2Real Reinforcement Learning for Soccer skills}

       \vspace{0.5cm}
            
    \vspace{0.5cm}
        By
        \\
        \vspace{0.3cm}
        Jonathan Spraggett 
       \\
        jonathan.spraggett@mail.utoronto.ca
        \\
        \vspace{0.8cm}
        Supervisor: Michael Guerzhoy
        \\
        April 14, 2023
        \\
         \vspace{7cm}

   \end{center}
\end{titlepage}
\vspace*{5cm}

\section*{\begin{center} Abstract\end{center} }

        This thesis work presents a more efficient and effective approach to training control-related tasks for humanoid robots using Reinforcement Learning (RL). The traditional RL methods are limited in adapting to real-world environments, complexity, and natural motions, but the proposed approach overcomes these limitations by using curriculum training and Adversarial Motion Priors (AMP) technique. The results show that the developed RL policies for kicking, walking, and jumping are more dynamic, and adaptive, and outperformed previous methods. However, the transfer of the learned policy from simulation to the real world was unsuccessful, highlighting the limitations of current RL methods in fully adapting to real-world scenarios. 
\newpage
\vspace*{3cm}
\section*{\begin{center} Acknowledgements\end{center}}

        I would like to express my sincere gratitude to Prof. Michael Guerzhoy for his invaluable guidance, insights, and suggestions throughout the course of this thesis. His constant support and encouragement have been instrumental in shaping my research skills and academic growth.
        \\\\\
        I would also like to extend my appreciation to the entire UTRA Robosoccer family and other UTRA members for their invaluable contributions to my research work. Their collective efforts and support have played an important role in providing me with the opportunity to pursue this thesis work.
        \\\\\
        Furthermore, I am deeply grateful to Shahryar Rajabzadeh for his exceptional guidance in navigating the PyBullet environment and openAI gym from his thesis research, and to Hoang Nam Nguyen for his invaluable assistance in generating simulation models from the CAD environment.
        \\\\\
        Thank you all for your support and encouragement throughout this journey.
   
\newpage
\newpage

\newpage
\tableofcontents
\newpage
\listoffigures
\newpage
\listoftables
\newpage
\pagenumbering{arabic}
\section{Introduction}
    The RoboCup Humanoid Soccer League is a competition where teams of humanoid robots play soccer against each other \cite{robo_challenge}. One of the most crucial aspects of a soccer match is control-related tasks such as movement and kicking. These tasks are vital as they determine the accuracy and speed of the robot's actions on the field. The robots need to have precise control over their movements, including the ability to make quick turns, changes of direction, and precise kicks. Furthermore, the robots must be able to coordinate their movements with their teammates to effectively move the ball down the field and score goals. Without accurate and efficient movement and kicking control, the robots would not be able to compete effectively in the RoboCup Humanoid Soccer League.
    \\\\\
    The environment in which the robots play is incredibly dynamic and constantly changing. They may face unexpected obstacles, such as uneven ground or contact with other players, which can cause disturbances in their movements and even lead to falls. This can result in a loss of possession of the ball, making it crucial for the robots to have a high level of resistance against external forces. This is an important aspect that must be considered when developing effective control-related tasks such as movement and kicking for the robots to perform well in the game.
    \\\\\
    Control of these robots can be challenging due to their complex structure, making basic skills such as movement and kicking difficult \cite{RL_biped}. Traditional techniques for control-related tasks such as kicking a soccer ball involve using static trajectories, which require manual determination of all motor positions \cite{kick}. This limits the robot's ability to react dynamically to the changing environment of a soccer match. A more adaptive technique for movement is the use of a sophisticated walking engine, which allows robots to obtain stable and smooth trajectories. However, creating and tuning a walking engine for custom hardware can take years and require a significant amount of knowledge and effort \cite{bit_walking}.
    \\\\\
    Reinforcement Learning (RL) is a popular approach for controlling humanoid robots. It trains models by rewarding desired behaviors \cite{russell2016artificial} but has limitations, such as the need for complex reward engineering to achieve natural and effective behaviors \cite{deepmimic}, as well as the reliance on simulated environments for training, which can lead to a mismatch between the simulated and physical world, making it difficult to transfer the model to a real-world environment \cite{dulac2019challenges}. To address these challenges, researchers have been exploring alternative approaches such as imitation learning. This technique allows the robot to imitate behaviors from a motion reference to produce more natural and transferable motions. However, imitation learning also has its own limitations, such as limiting the robot's ability to develop versatile and diverse behaviors to fulfill auxiliary task objectives, as well as an increased development effort for selecting and tuning motion references \cite{deepmimic}. As a result of these challenges, no team within the RoboCup Humanoid Soccer league has yet been able to successfully apply RL to control-related tasks on a physical robot \cite{rhoban_team}.
    \\\\\
    This thesis aims to develop a more efficient and effective approach for training control-related tasks for humanoid robots that can overcome the limitations of traditional Reinforcement Learning methods and adapt to real-world environments. The proposed approach combines deep reinforcement learning with the Proximal Policy Optimization (PPO) algorithm to train a neural network in a simulated environment. PPO is a widely-used and effective method for training RL agents, which can improve the stability and reduce the programming effort to perform the motion \cite{russell2016artificial}. Additionally, the thesis implements Adversarial Motion Priors (AMP) technique to improve the performance of the robot. AMP is a technique that combines imitation learning and Generative adversarial network (GAN) style training to enable the robot to imitate behavior from a reference motion dataset, allowing for greater flexibility and freedom to develop new behaviors to achieve task objectives while requiring less complex reward functions. This reduces the likelihood of producing undesirable movements and makes the model more reliable \cite{amp}. The trained model is fine-tuned and deployed in a real-world setting to achieve optimal performance. To address any discrepancies between the simulated training environment and the actual conditions, the method employs domain randomization, which enhances the model's ability to generalize by introducing diverse variations in the training environment \cite{domain}. 
    \\\\\
    This proposed approach will enable a humanoid robot to dynamically react to the environment of a soccer match and make more complex decisions, including team play. Furthermore, this research lays the foundation for future Reinforcement Learning projects that utilize Adversarial Motion Priors to transfer complex skills learned in simulation to physical robots.

\newpage
\section{Fundamentals}
    \subsection{RoboCup}
\begin{figure}[h]
        \centering
        \includegraphics{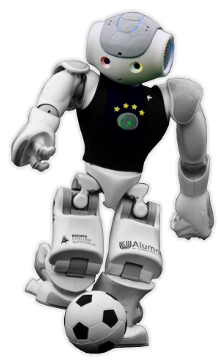}
        \caption{Standard Platform League Robotic platform \cite{spl}}
        \label{fig:spl}
        \end{figure}

\begin{figure}[h]
        \centering
        \includegraphics[width=150mm]{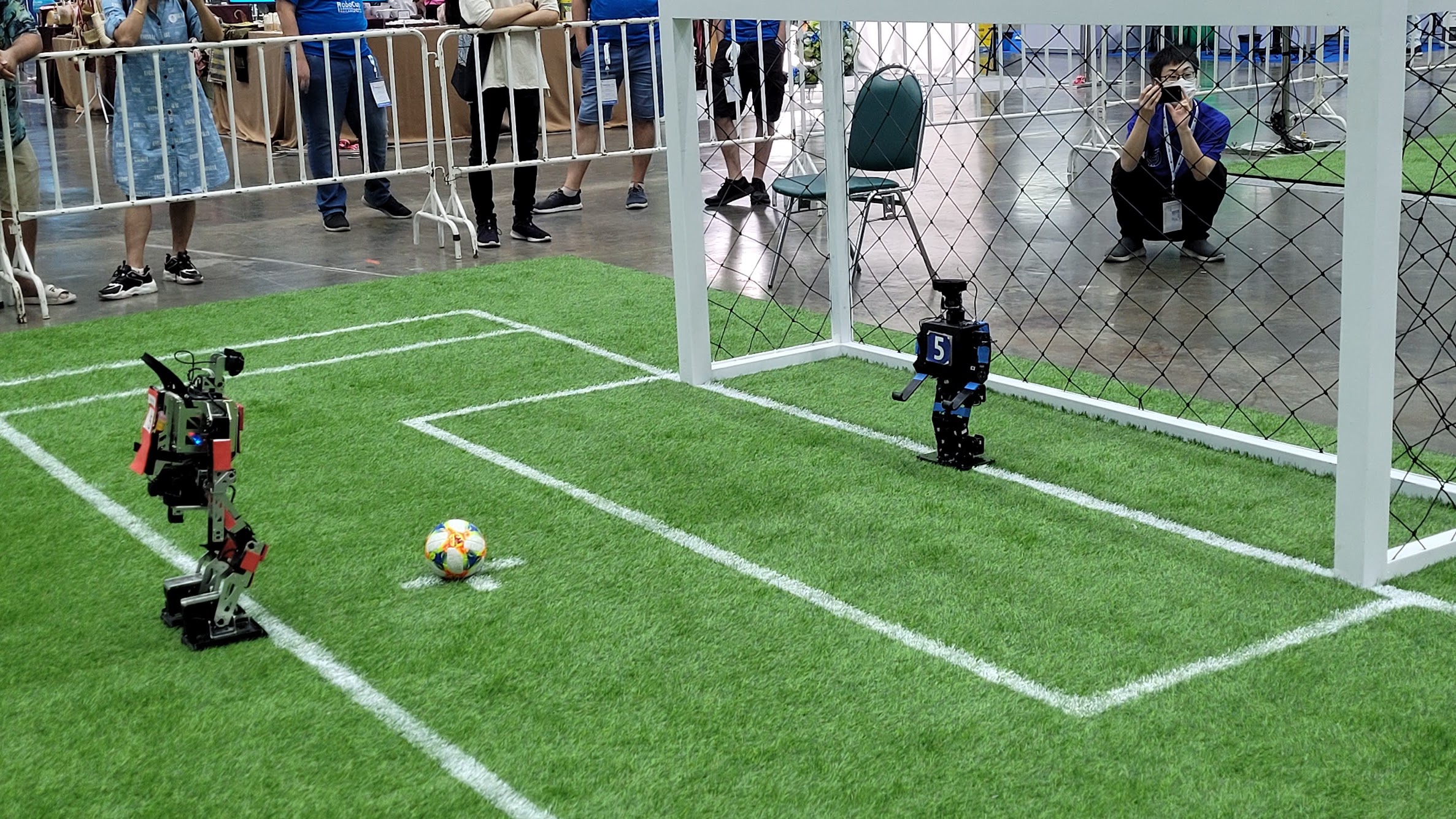}
        \caption{Two robots in the Humanoid Kid Size League \cite{robot}}
        \label{fig:game}
        \end{figure}

        RoboCup is a prestigious international robotics competition that aims to advance research and education through the use of soccer games. Established in 1997, its ultimate goal is to create a team of autonomous humanoid robots that can defeat the 2050 FIFA World Cup champion team in a match \cite{robo_challenge}. The competition is divided into various leagues, each of which poses a unique challenge and requires a distinct type of robot. These leagues provide a platform for researchers and students to showcase their advancements in the field of robotics and artificial intelligence \cite{robocup}.
        \\\\\
        The Robocup simulation league is one of the original leagues created, which utilizes simulated environments to simulate soccer matches played by teams of autonomous robots \cite{robocup}. This league is often at the forefront of cutting-edge research in software advancements, as there are fewer hardware limitations. The league has been a pioneer in the use of reinforcement learning in Robocup.
        \\\\\
        Another important league in Robocup is the Standard Platform League, where teams use the standardized humanoid robot, the NAO robot by SoftBank, to play soccer (refer to Figure \ref{fig:spl})  \cite{robocup}. This league is also at the forefront of research in software advancements for physical robots and has been instrumental in the early use of reinforcement learning on physical robots in Robocup.
        \\\\\
        The UTRA RoboSoccer team was established in 2017 with the goal of competing in the RoboCup Kid-size Humanoid League. This league is similar to the Standard Platform League. However, the Kid-size Humanoid League places an emphasis on custom-designed humanoid robotic platforms, developed independently by each team. The structure of the robots is limited to resemble a humanoid with two arms, legs, and one head. Additionally, certain body segment dimensions are constrained to ensure that the robot maintains a humanoid appearance. The sensors used on the robots are restricted to those found on humans, such as cameras, inertial measurement unit (IMU), and force sensors. Figure \ref{fig:game} illustrates an example of these custom humanoid robots during a game \cite{robocup_rules}.
        
    \newpage
    \subsection{Robot Platform}
        
        The robotic platform used in this thesis is the Bez platform, as referenced in \cite{platform} and seen in figure  \ref{fig:bez} below. The design of the Bez platform was modified from the DARwIn-OP platform, as detailed in \cite{darwin}, to optimize for 3D printing and ease of assembly. The robot possesses 18 degrees of freedom, including 6 per leg, 2 per arm, and 2 in the head. The robot's physical specifications include a height of 50cm and a weight of 2.3kg. An inertial measurement unit (IMU) is located in the center of the front chest plate, which is used to measure linear acceleration and angular velocities for determining the robot's orientation. The camera utilized on the robot is a Logitech C920 HD camera with a resolution of 1920x1080 pixels. The actuators employed on the robot are Dynamixel MX28 and AX12 servos, which utilize position control to compute applied torque on the servo by means of internal PID controllers based on a target position.
        \begin{figure}[h]
        \centering
        \includegraphics[width=75mm]{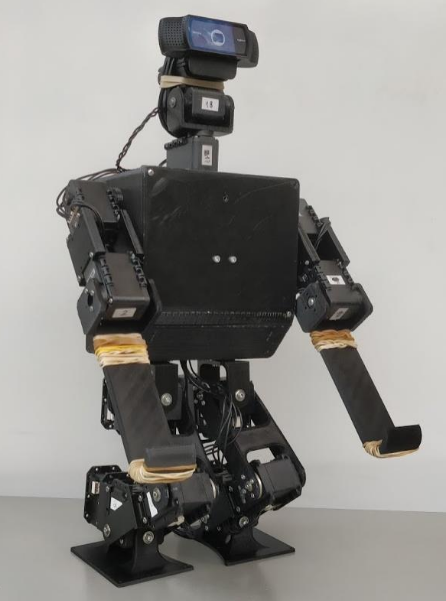}
        \caption{Bez platform \cite{platform}}
        \label{fig:bez}
        \end{figure}
        \\\\\
        The Bez platform, as described in reference \cite{platform}, is the robotic platform used in this thesis. The design of the Bez platform was modified from the DARwIn-OP platform \cite{darwin}, in order to optimize for 3D printing and ease of assembly. The robot features 18 degrees of freedom, including 6 per leg, 2 per arm, and 2 in the head. Its physical specifications include a height of 50cm and a weight of 2.3kg. An inertial measurement unit (IMU) is located in the center of the front chest plate, which is used to measure linear acceleration and angular velocities for determining the robot's orientation. The camera used on the robot is a Logitech C920 HD camera with a resolution of 1920x1080 pixels. The actuators employed on the robot are Dynamixel MX28 and AX12 servos, which use position control to compute the applied torque on the servo by means of internal PID controllers based on a target position.
        \\\\\
        The primary computing unit used in this platform is the Nvidia Jetson Tx2, which executes the entire code stack, including the walking engine and computer vision modules. The STM32F446RE microcontroller manages the communication and sensor and actuator data.

    \newpage
    \subsection{Reinforcement Learning}
        Reinforcement Learning (RL) is a machine learning technique that trains agents how to learn and make decisions based on the feedback they receive. The agent's goal is to learn a policy that maximizes the cumulative reward over time. RL is used to solve a wide range of problems, including control, navigation, and game-playing.
        \\\\\
        The basics of the RL approach require several components: the agent, the environment, and the reward signal. The agent interacts with the environment, taking action and receiving rewards or penalties. 
        The environment is the system that the agent interacts with. The reward signal is the feedback provided to the agent indicating its performance. The goal is to learn a policy, a set of rules for how the agent should behave, that will result in the highest cumulative reward over time \cite{russell2016artificial}.
        \\\\\
        The training follows the following process:
       \begin{enumerate}
        \item Observation: The agent observes the current state of the environment. This state can be represented as a set of variables that describe the environment.
    
        \item Selection: The agent selects an action to perform based on its current policy. The policy is a set of rules that determines which action the agent should take in a given state.
        
        \item Execution: The agent performs the selected action and the environment transitions to a new state.
        
        \item Feedback: The agent receives a reward signal from the environment as a result of its action. The reward signal is a scalar value that indicates how well the agent is performing.
        
        \item Learning: The agent updates its policy based on the observed state, action, reward, and next state. This step is often referred to as the learning step.
        
        \item Iteration: The agent repeats steps 1 to 5 until it reaches a stopping condition. The stopping condition can be a maximum number of steps, a threshold reward, or a specific state.
        
        \end{enumerate}

     
    \subsection{Proximal Policy Optimization}
        Proximal Policy Optimization (PPO) is a technique used in Reinforcement Learning (RL) to help improve the stability and consistency of policy updates. The idea behind PPO is that it is better to make small, incremental changes to the current policy, rather than make large changes. The "clip" function achieves this functionally by limiting the change in the policy.
        This prevents the agent from getting stuck in a decision-making loop and ensures that the policy is updated in a stable and consistent way \cite{ppo}.
                \\\\\
        There are two main components: the actor and the critic. The actor decides what action to take based on the current state, while the critic evaluates the performance of the current policy and decides whether to accept or discard and revert to the old policy.
        \\\\\
        The algorithm follows the following process:
        \begin{enumerate}
    
        \item The agent collects data such as states, actions, and rewards of the agent from the environment. 
        
        \item The agent updates its policy by taking small steps in the direction of the policy that should return the highest rewards.
        
        \item The agent uses the updated policy to generate more data from the environment.
        
        \item The agent uses this new data to continue updating its policy.
        
        \item The agent repeats steps 2-4 for a number of iterations.
        
        \item After, the agent evaluates the performance of the updated policy by comparing it to the previous policy.
        
        \item If the updated policy performs better, it is accepted and the algorithm continues. If the updated policy performs worse, it is discarded and the algorithm goes back to the previous policy.
        
        \item The agent continues to repeat the process until it reaches a stopping condition.
        \end{enumerate}
        
        
    \subsection{Imitation Learning}
    Imitation learning is a machine learning technique that leverages the expertise of a human or an expert agent to solve complex control problems. It is a combination of reinforcement learning and motion tracking, where the agent learns by observing and mimicking an expert's action \cite{schaal1996learning}.
    \\\\\
    In one common implementation of imitation learning, the agent is trained using supervised learning to predict the expert's actions given the current state of the environment. The learned model is then used to guide the agent's own actions \cite{deepmimic}.
    \\\\\
    Another approach is inverse reinforcement learning, where the agent infers the expert's underlying reward function from their behavior and uses this information to train its own control policy. This approach allows the agent to learn more complex and nuanced behaviors that may not be easily modeled through traditional reward functions \cite{irl}.

    \subsection{Advanced Motion Prior}
     The Adversarial Motion Priors (AMP) approach combines imitation learning and Generative adversarial network (GAN) training to enable robots to perform complex tasks while imitating behaviors from large, unstructured datasets. The technique leverages adversarial techniques to allow robots to learn from reference motion datasets.
    \\\\\
    The AMP approach involves training a discriminator network to differentiate between the robot's movements generated by a control policy and the movements depicted in the reference dataset. The discriminator serves as the style reward for training the control policy, enabling the robot to imitate the behaviors in the demonstration data more naturally.
    \\\\\
    By combining the strengths of imitation learning and GAN-style training, the AMP approach allows robots to produce more sophisticated and realistic movements while retaining the ability to generalize and adapt to new situations. As a result, the robots can perform high-level tasks more effectively, and the learned style reward leads to more natural and human-like behaviors. \cite{amp}
    
      
\newpage
\section{Related Work}
    \subsection{Classical Approach}
        In the early days of Robocup, control was achieved through basic methods such as rule-based systems and reactive control. Although these techniques were effective, they were limited in terms of sophistication compared to current methods. Static keyframe animations have become a widely used approach in Robocup, even today, for specific actions like kicking and getting up \cite{survey}. This method involves recording a series of motor positions and timings that are then interpolated to generate the movement. The programming process does not require extensive knowledge of the robot's dynamics and offers a low-tech solution that is relatively easy to implement. However, recording the keyframes can be a time-consuming task, and fine-tuning the positions accurately can be challenging. Additionally, the static nature of keyframe animations means that the robot cannot react to its surroundings, which can limit its effectiveness in certain scenarios \cite{kick}. 
        \\\\\
        In addition to static keyframe animations, control engines are another widely used approach in Robocup for tasks such as walking and kicking \cite{survey}. These engines offer a sophisticated solution to achieve stable and smooth movements. Examples include the real-time adapting dynamic kick engine [kick engine paper] and the easy-to-use omnidirectional walk controller based on parameterized splines and closed-loop stabilization. These controllers use feedback from IMU and servo data to maintain stability. [biped paper]. However, creating a custom walk engine for hardware can pose a significant challenge and require years of effort, making it difficult for new teams to enter the RoboCup Humanoid League due to the need for both functional hardware and a comprehensive software solution \cite{robo_challenge_2}.     
        \newpage
    \subsection{Reinforcement Learning}
        Reinforcement Learning (RL) has become a popular method for solving control problems in recent years \cite{rl_survey}. Instead of using manual control methods, RL uses an automated approach to make control decisions based on a set goal. The progress in computing technology has made it possible to use RL in complex systems without a lot of technical knowledge of the domain \cite{shah_cpu}.
        \\\\\
        An example of the successful application of RL can be seen in Nvidia's Isaac gym simulated environment, where various robotic applications of locomotion and manipulation were demonstrated to be effectively controlled by RL, both in simulation and in real-world scenarios \cite{makoviychuk2021isaac}. One example is the use of RL to train a quadrupedal robot, Anymal, to walk on uneven terrain. The training was conducted using the PPO algorithm and designed a reward system to reward forward motion and avoid big changes in velocity. They also introduced a novel curriculum approach to improve performance and train the robot for more difficult tasks. As a result, the robot was able to learn how to walk well and the method was applied in real-world situations \cite{rudin2021learning}.
        \\\\\
        Within the RoboCup community, RL has been gaining increased attention. For instance, in 2014 a team in the standard platform utilized RL to develop a fuzzy controller for an NAO robot, enabling it to dribble a ball with high accuracy \cite{fuzzy}. In 2021, a simulation league team applied PPO and RL to create a multi-directional kick that was significantly more adaptable and dynamic compared to previous state-of-the-art methods \cite{toes}. However, these approaches were limited to standard robotic platforms like the NAO or solely applied within simulation leagues.  
        \\\\\
        Despite its successes, RL-based control is not without limitations. These include undesirable behaviors and difficulties in transferring learned policies to real-world environments. The policies learned through RL may result in unnatural, jerky movements that may not be suitable for practical deployment and may require manual design constraints to regulate behavior. Furthermore, these methods are often specific to a particular task, requiring substantial effort to tune for each new skill and leading to complex reward functions. Additionally, learned motions may not be effective in real-world scenarios due to inaccuracies in simulation \cite{complex}, \cite{deepmimic}.
        
    \subsection{Imitation Learning}
        Imitation Learning and Domain Randomization are advanced techniques that address common challenges in Reinforcement Learning. Imitation Learning allows robots to imitate motion data and accelerate learning faster than with traditional reinforcement learning because it observes expert actions and adapts its own behavior \cite{schaal1996learning}, while Domain Randomization enables better transfer of learned skills to real-world environments.
        \\\\\
        For instance, DeepMimic used Imitation Learning and PPO to develop complex motions such as throwing, running, and kicking, with improved performance compared to traditional reinforcement learning [deepmimic]. In 2022, a team in the RoboCup community used Imitation Learning, Domain Randomization, and PPO to train a Darwin robot to walk effectively and reliably in uneven terrain, faster than the previous state-of-the-art \cite{lorm}.
        \\\\\
        Motion tracking combined with reinforcement learning is effective for complex and dynamic skills in simulated domains. However, Imitation Learning has limitations, such as limiting the robot's ability to develop versatile and diverse behaviors and increasing development efforts for selecting and tuning motion references \cite{complex}. Which restricts how it can be applied to certain domains.
        \\\\\
        The use of Domain Randomization enhances the ability of a model to generalize and transfer learned skills to real-world conditions. This method randomly introduces diverse variations in the training environment to account for any discrepancies between simulation and actual conditions \cite{domain}. In recent years, this approach has been shown to be effective, with noteworthy results achieved in the case of walking on a two-legged robot through the use of Domain Randomization \cite{RL_biped}.        
        \subsection{Adversarial Imitation Learning}
        Adversarial Imitation Learning and Adversarial Motion Priors are cutting-edge methods that tackle common problems in imitation learning. AMP combines these techniques with additional objectives to enable robots to perform complex tasks while imitating movements from diverse datasets. The use of adversarial methods results in policies that align with the distribution of movements in the dataset, providing greater versatility and adaptability. This advanced system can produce motions of exceptional quality, comparable to the best techniques available, and can work effectively for a range of characters and tasks without the need for complicated motion planning. With AMP, robots can move like humans without the risk of making mistakes, making the system more dependable and efficient \cite{amp}.
        \\\\\
        AMP has been successfully applied to complex tasks, such as training a quadrupedal robot to perform natural and energy-efficient locomotion using a less complex reward function \cite{complex}. Additionally, Multi-AMP was developed to train a single controller for multiple skills and styles of motion, demonstrating improved performance over previous models. These systems were then transferred to a physical platform and performed better than previous models\cite{multi_amp}.
        \\\\\
        However, AMP is still experimental and has not been applied in the RoboCup competition. Despite this, the potential for AMP to lead to more natural, physically plausible, and energy-efficient behaviors for legged robots is promising \cite{rhoban_team}.

\newpage
\section{Simulation Environments}
    \subsection{Nvidia Isaac Gym}
    
        Nvidia Isaac Gym was chosen as the simulation environment for training the policy due to its efficient GPU parallelization capability, allowing thousands of robots to learn simultaneously on a single GPU (refer to figure \ref{fig:sim}), speeding up training time by 2-3 orders of magnitude compared to CPU-bound simulations like PyBullet and MuJoCo \cite{makoviychuk2021isaac}.
        \\\\\
        The simulation runs at 120Hz to mimic the real-world control loop of Bez's sensors, motors, and control frequency. Bez is initially placed on the ground in a ready position, and a soccer ball is added based on the task being trained. The environment's physics settings, such as friction and gravity, were left as default for optimal results, as seen in the successful training of Anymal, a quadrupedal robot, to walk \cite{rudin2021learning}. The simulator runs in a headless mode to save resources during training, with no graphical interface.

        \begin{figure}[h]
            \centering
            \includegraphics[width=150mm,scale=1]{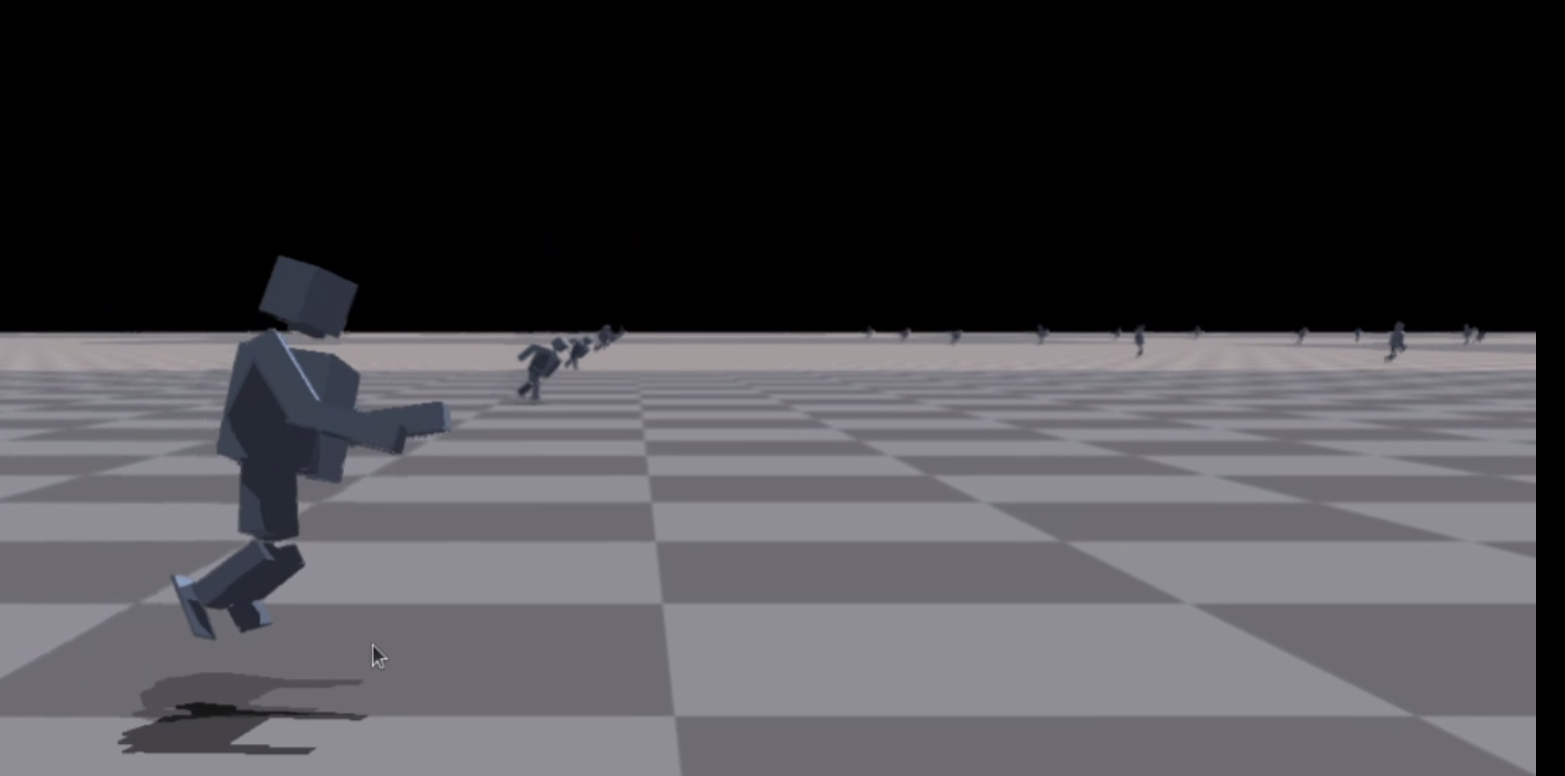}
            \caption{Bez URDF model in Isaac gym Learning how to jump with 4096 instances running simultaneously}
            \label{fig:sim}
            \end{figure}
        
    \subsection{Webots}
        The simulation environment selected for the initial transfer of control policies from a simulated to higher fidelity simulated environment is Webots. This choice is informed by the fact that Webots is the official simulation environment used in the Humanoid League Virtual Season (HLVS), a virtual competition for verifying and testing robot code prior to the physical competition. In order to ensure the transferability of the approach to the HLVS, the contact properties and other simulation parameters, such as the simulation step duration, are derived from the HLVS environment, as detailed in \cite{robo_challenge}
    \begin{figure}[h]
            \centering
            \includegraphics[width=150mm,scale=1]{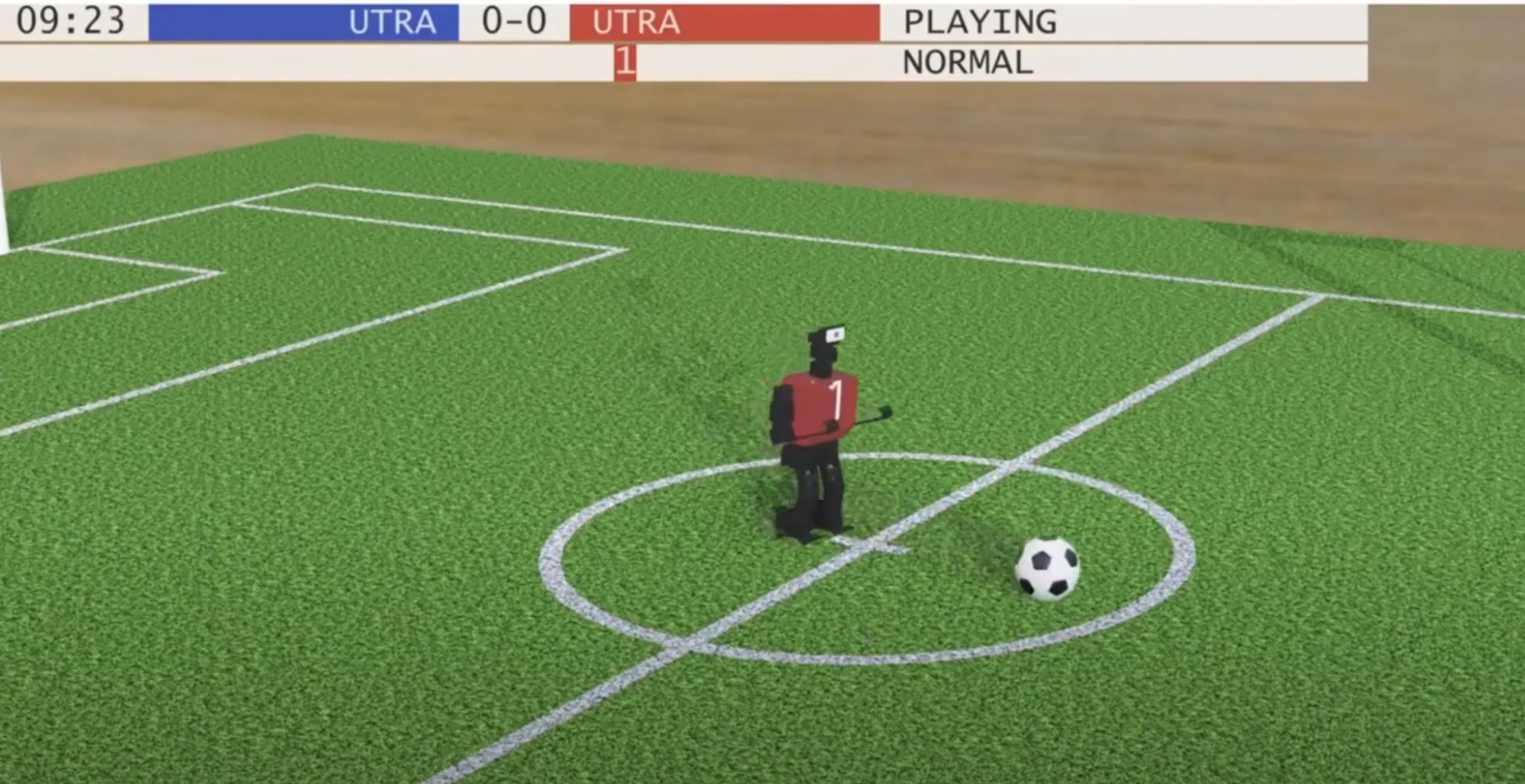}
            \caption{Bez PROTO model in Webots}
            \label{fig:sim2}
    \end{figure}

            \newpage
    \subsection{Robot Models}
\begin{figure}[h]
            \centering
            \includegraphics[width=75mm,scale=0.5]{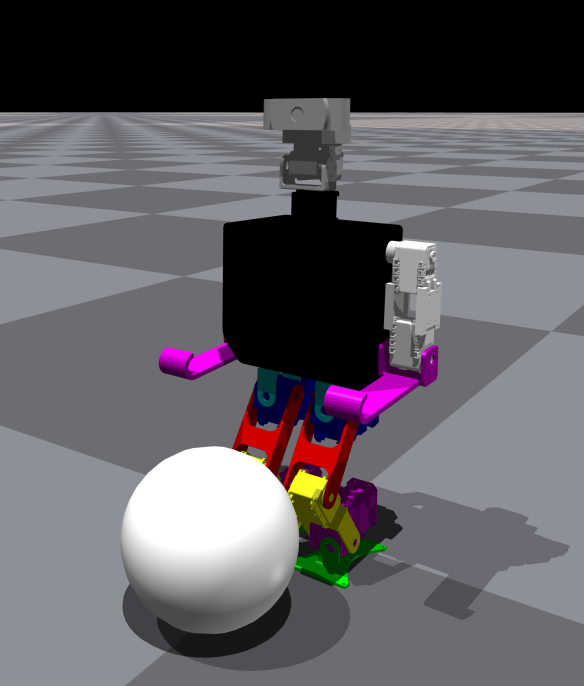}
            \caption{Bez URDF model in Isaac gym with a soccer ball in front}
            \label{fig:bez_sim}
        \end{figure}

        The robot model was generated based on the CAD files of the second iteration of the Bez robot designed in AutoDesk Fusion 360. It was manually created in the Unified Robot Description Format (URDF) and the collision meshes were approximated with the smallest bounding boxes. The model can be viewed in figure \ref{fig:bez_sim} below
        \\\\\
        
        Joint limits were imposed in the simulation environment to comply with RoboCup's human-like motion rule \cite{robocup_rules}. These limits were meant to ensure that the resulting trajectory is within the capabilities of a human. A software-based solution is implemented to bring any joint that goes beyond reasonable human limits back into acceptable ranges by overriding the velocity controller.
        \\\\\
        The robot model, accessible at \cite{bez}, uses bounding boxes to define the robot's boundaries for collision detection. Each URDF link is approximated as a box of similar dimensions, and the collision between parent and child links is disabled in the simulation environment.
        \\\\\
        To comply with RoboCup regulations, the range of motion for each joint while standing was estimated individually. The constraint values for the left leg and arm are the same, and no mirroring was required.
        \\\\\
        This URDF was adapted to MuJoCo Modeling XML File (MJCF) to be used in the AMP version of the training. The key differences were changing the motor characteristics and orientation to be better aligned with the motion data and the density of physical components to ensure a more reliable motion.
        \\\\\
        Simulation environment Webots does not accept URDF or MJCF and requires converting the URDF into PROTO type file. Manual adjustments were made to align body parts, sensors, and bounding boxes
        
    \subsection{Motor Control}
    The robot uses position control to replicate the current control method and to evaluate if simpler control methods, using budget-friendly servo motors that may lack velocity and torque control, can be effective. The servo motors used are AX-12 and MX-28 Dynamixel, with specifications in tables \ref{table:mx28} and \ref{table:ax12} below.

    \begin{table}[h!]
    \caption{MX28 motor characteristics \cite{mx28}}
    \centering
    \begin{tabular} {||c | c||} 
     \hline
     \multicolumn{2}{||c||}{MX28} \\
     \hline
     Parameter & Value\\ [0.5ex] 
     \hline
     Stall torque & 2.5 [N.m] \\ 
     \hline
     No load speed & 55 [rev/min]  \\
     \hline 
     Sensor Resolution & 0.33 [°]  \\
     \hline  
    
    \end{tabular}
    
    \label{table:mx28}
    \end{table}

    \begin{table}[h!]
    \caption{AX12 motor characteristics \cite{mx12}}
    \centering
    \begin{tabular} {||c | c||} 
     \hline
     \multicolumn{2}{||c||}{AX12} \\
     \hline
     Parameter & Value\\ [0.5ex] 
     \hline
     Stall torque & 1.5 [N.m] \\ 
     \hline
     No load speed & 59 [rev/min]  \\
     \hline 
     Sensor Resolution & 0.29 [°]  \\
     \hline  
    
    \end{tabular}
    
    \label{table:ax12}
    \end{table}
    \subsection{Inertial Measurement Unit (IMU)}
    IMUs, or Inertial Measurement Units, are crucial components in robotics, allowing robots to maintain stability, detect falls and track orientation. The calculation of IMU signals starts with determining the linear velocity and angular velocity of a link in the world's frame. The linear acceleration can then be calculated using the equation below: 
    \\\\\
    $linear \; acceleration= (current\; linear\; velocity - previous\;linear\; velocity) / control\; loop \;time\; step$
    \\\\\
    In order to accurately simulate the real-world behavior of an IMU, the location of the IMU in the torso, as well as the inherent noise and drift due to temperature, must be taken into consideration. This can be achieved by modeling a Gaussian random variable based on the STMicroelectronics LSM6DSOX \cite{stm}. Table \ref{table:imu_noise} below shows the noise standard deviation added to the sensor.
    
    \begin{table}[h!]
    \caption{IMU Noise Percentages.}
    \centering
    \begin{tabular} {||c | c||} 
     \hline
     Sensor & Noise (\% of max boundary)\\ [0.5ex] 
     \hline
     Accelerometer & 0.203\% \\ 
     \hline
     Gyro & 0.804\%  \\
     \hline 
    
    \end{tabular}
    
    \label{table:imu_noise}
    \end{table}

    \subsection{Foot-Surface Contact Detection}
\begin{figure}[h]
                \centering
                \includegraphics[width=75mm,scale=0.5]{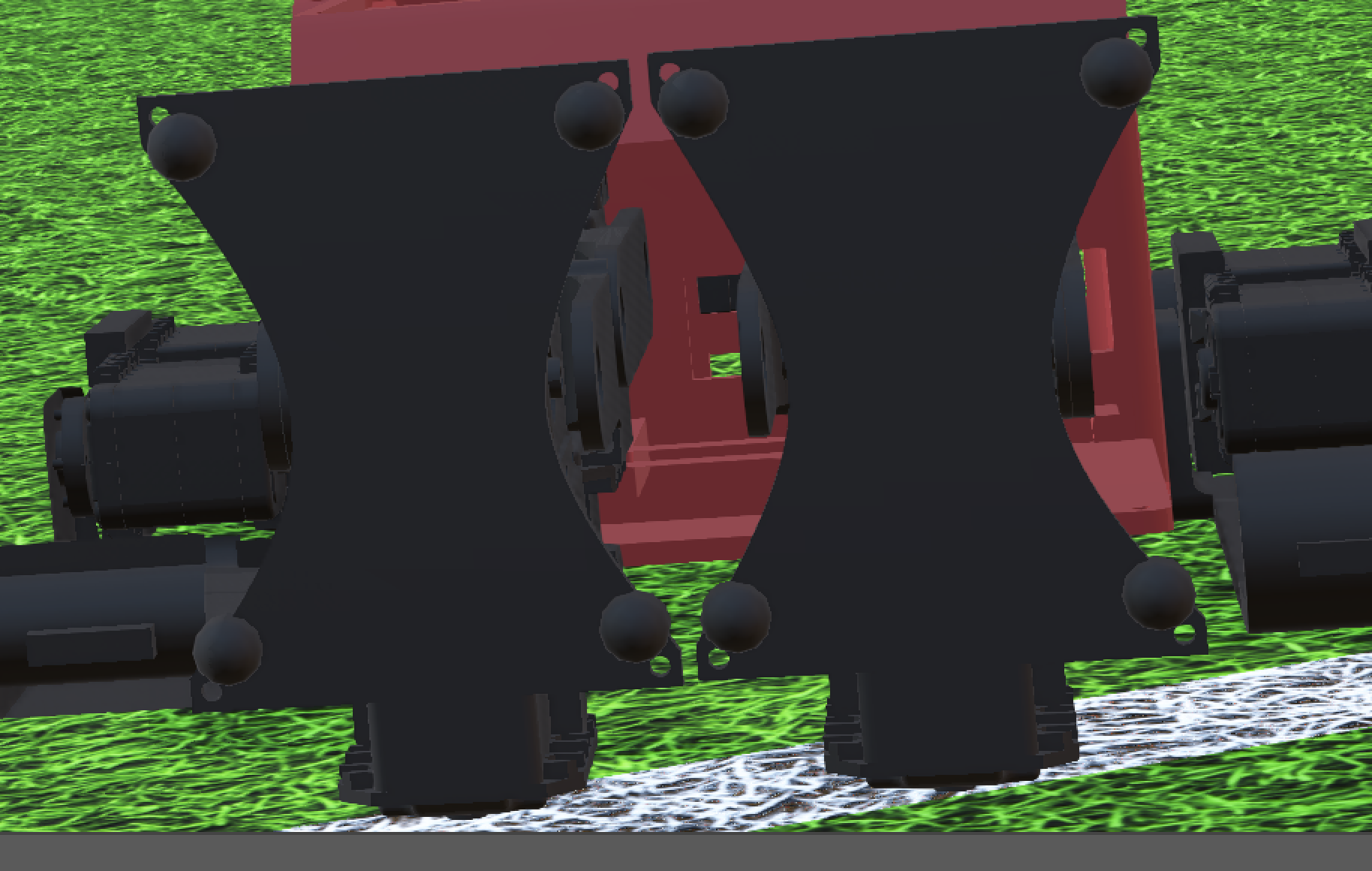}
                \caption{Bez URDF model with foot pressure sensors attached}
                \label{fig:feet}
                \end{figure}

            Foot sensors play a crucial role in ensuring the stability and control of robots. To demonstrate the importance of foot sensors, the winning team of the RoboCup competition from 2016 to 2019 improved the stability of their robot by utilizing force measurements from the four corners of their robot \cite{rhoban_team}. Bez URDF was adapted to add foot force sensors but faced challenges in accounting for friction and differentiating it from normal forces in real life. As a result, choosing to detect contacts at each quadrant of each foot, resulting in a vector of 8 binary values, with 4 values for each foot. The function checks if the four corners of each foot are in contact with the ground. The footplates, when viewed from above, have indices labeled as 4-5 for the left foot and 0-1 for the right foot. This approach returns an array of 8 contact points on both feet, with a value of 1 indicating that a point is touching the ground and -1 otherwise. This approach is less sensitive than using continuous force sensors but still provides valuable data for stabilization. Figure \ref{fig:feet} shows foot sensors attached to Bez feet.
                
    \newpage
\section{Training Process}
    \subsection{Implementation }
        Revised: The training process uses the RL Games software library, which leverages GPU parallelization and runs entirely on the GPU for increased training speeds, and has access to various popular RL algorithms \cite{makoviychuk2021isaac}.
        \\\\\
        The training is divided into episodes, each representing a specific task (e.g. walking, kicking, jumping) to be completed within a given time frame. Bez starts in a ready state and, depending on the task, a ball may appear in front. The environment generates an observation state vector, which is passed to the policy to produce actions that are applied to the robot's motors.
        \\\\\
        To terminate an episode, the simulation environment tracks the robot's height and ends the episode when its upper torso falls below 27.5 cm, with a reward of zero.
        \\\\\
        Table \ref{table:computer} shows the computer specification used to train the policies.

        \begin{table}[h!]
        \caption{Training Server Specification}
        \centering
        \begin{tabular} {||c | c ||} 
         \hline
        Computer Component  & Name \\
         \hline
         CPU & AMD 3900X \\
         \hline
         GPU & Nvidia 3090   \\
         \hline
         RAM & 64 GB Quad Channel, 3600 MT/s \\
        \hline
        
        \end{tabular}   
        \label{table:computer}
        \end{table}

    \subsection{Network Architecture}

\begin{figure}[h]
            \centering
            \includegraphics[width=85mm ]{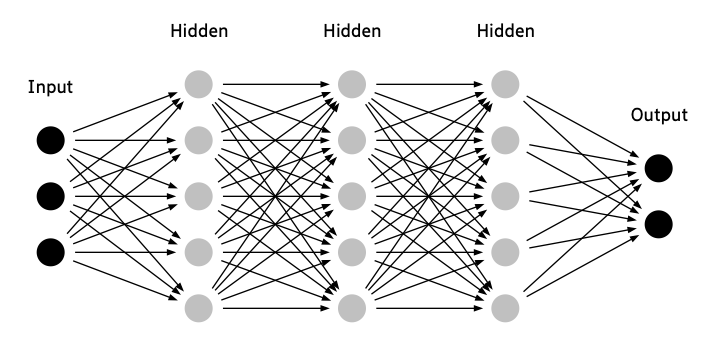}
    
            \caption{The actor-network consists of three fully connected hidden layers with 400x200x100 neurons for each layer respectively. The ReLU activation function is used.}
            \label{fig:nn}
        \end{figure}

        This thesis adopts the Actor-Critic variant of Proximal Policy Optimization (PPO) and employs two neural networks to carry out the algorithm's operations. The first network is dedicated to the policy function, while the second network is used to evaluate the actions generated by the policy.
        \\\\\
        The architecture of each of the networks consists of three fully connected hidden layers, all of which utilize the Rectified Linear Unit (ReLU) activation function. The output layer of the value function is linear with a single output neuron, while the policy function has an output layer with a number of neurons that corresponds to the number of available actions. Figure \ref{fig:nn} shows a visualization of the neural network architecture for the actor-network.
        \\\\\

        The selection of hyperparameters seen in table \ref{table:hp} and neural network architectures was based on the results obtained in the \cite{rudin2021learning} experiments, as they demonstrated remarkable outcomes.
        \begin{table}[h!] 
        \caption{Hyperparameter for Neural network} 
        \centering
        \begin{tabular} {||c | c||} 
         \hline
         Parameter  & Value   \\
         \hline
         Policy Network & 400x200x100, fully connected Activation Function: ReLU \\
         \hline
         Value Function Network  & 400x200x100, fully connected Activation Function: ReLU \\
         \hline
         Minibatch Size  & 32768 \\
        \hline
        Number of mini epochs  & 5 \\
        \hline
        GAE-$\lambda$  & 0.99 \\
        \hline
        Discount Factor $\gamma$  & 0.95 \\
        \hline
        PPO Clip Range   & 0.2 \\
        \hline
        Learning Rate  & 3e-4 \\
        \hline
        Value Function Loss Coefficient  &  0.001 \\
        \hline
        Entropy Coefficient  &  0.0 \\
        \hline
        
        \end{tabular}
          
        \label{table:hp}
        \end{table}
    \newpage
    \subsection{Observation}
    The observation state vector is the key factor that drives the policy network and influences its decision-making process. It encompasses various elements such as the current state of the task being executed, the robot's status obtained from various sensor readings, and any supplementary environmental information. The selection of relevant sensor data for the observation network was adapted from \cite{rudin2021learning} and \cite{lorm}. 
    \\\\\
    The sensors installed on a humanoid robot have limited capabilities and do not have access to external sources. Hence, the robot must depend on its observation of the readings from these sensors to determine its state. The specific components of the observation utilized in this study are outlined in table \ref{table:obs}.
    \\\\\
    \begin{table}[h!]
    \caption{Observation State vector}
        \centering
        \begin{tabular} {||c | c||} 
         \hline
         Parameter  & Dimension \\
     \hline
     Motor Joint Positions &  18x1   \\
     \hline
     Motor Joint Velocity & 18x1  \\
     \hline
     Angular Velocities  &  3x1  \\
     \hline
     Linear Velocities & 3x1 \\
     \hline
     2D Orientation Unit vector from robot to goal vector  & 2x1  \\
     \hline
      Ball position (Only for Kicking task) &  2x1  \\
     \hline
     Feet contact with the ground (Only when enabled )&   8x1   \\
     \hline
        
        \end{tabular}
            
        \label{table:obs}
        \end{table}
    \\\\\
     To better understand the current state of a robot with position control motors, providing the policy with the current joint positions and velocities is crucial. This information allows us to comprehend the robot's current configuration and the direction in which each motor is moving. Additionally, the angular and linear velocities provide insights into the robot's current motion and trajectory.
    \\\\\
    To further enhance the robot's navigation capabilities, the 2D orientation unit vector plays a critical role by indicating the direction in which the robot should proceed. By considering all these parameters, more accurate and effective policies for controlling the robot's movement and achieving desired outcomes. 
    \subsection{Action}
    The action space in this thesis is defined as an 18-dimensional vector that represents the desired position targets for each of the motors in the robot's design. To ensure that the actions are evenly distributed across the outputs of the policy's Neural Network, they are scaled within the range of $[-\pi, \pi]$.
    \begin{table}[h!]
    \caption{Action State vector}  
        \centering
        \begin{tabular} {||c | c||} 
         \hline
         Parameter  & Dimension \\
     \hline
     Motor Joint Positions &  18x1   \\
     \hline

        \end{tabular}
          
        \label{table:obs}
        \end{table}
    \subsection{Rewards}
        The reward function is a crucial component in deep reinforcement learning, as it determines which behaviors are incentivized and which ones should be avoided. It serves as guidance and provides feedback to the agent, shaping its learning process. A well-designed reward function aligns with the task objectives and provides meaningful feedback.
    \subsection{Sim2Real}
        Once a policy was trained, it underwent retraining using domain randomization that incorporated Gaussian noise to all simulation parameters, including physics, gravity, and friction, as well as all sensor measurements. This technique was utilized to enhance the generalizability and adaptability of the policy's search space. Subsequently, the policy was transferred to Webots, where it interacted with the sensor output and motor inputs via the Robot Operating System (ROS).
    \newpage
\section{Skill Specifications}
    \subsection{Kick}
        \subsubsection{Description}
            The kicking skill is different from the others such that it requires Bez to interact with a game piece and manipulate it toward scoring a goal. At the start of every episode, a ball will spawn in front of Bez. Bez must extend his leg in a way to deliver the maximum amount of force on the ball without losing stability. Bez must also be accurate and cannot stray from the designated angle. 
        \subsubsection{Reward Engineering}
            The task reward at a high-level incentives Bez to approach the ball quickly by assigning the $VelocityForwardReward$, which is calculated based on the relative distance between Bez and the ball. The reward formula is as follows:
            
            \[DistanceToBall = BallPosition - BezPosition\]
            \[VelocityForwardReward =\frac{DistanceToBall}{\|DistanceToBall \|} \cdot IMULinearVelocityXY \]
            
            Additionally, the reward also prioritizes maximizing the forward velocity of the ball with $BallVelocityForwardReward$, to ensure a fast and impactful kick toward the goal. The formula is:
            
            \[DistanceToGoal = GoalPosition - BallPosition\]
            \[BallVelocityForwardReward =\frac{DistanceToGoal}{\|DistanceToGoal \|} \cdot BallVelocityXY \]
            
            These two components are then combined to form the final reward function:
            
           \[Reward = VelocityForwardReward + BallVelocityForwardReward\]
            
            The reward function aims to balance Bez's speed in reaching the ball and the velocity of the kick toward the goal.
        
        \subsubsection{Curriculum}
\begin{figure}[h]
            \centering
            \includegraphics[width=150mm,scale=1]{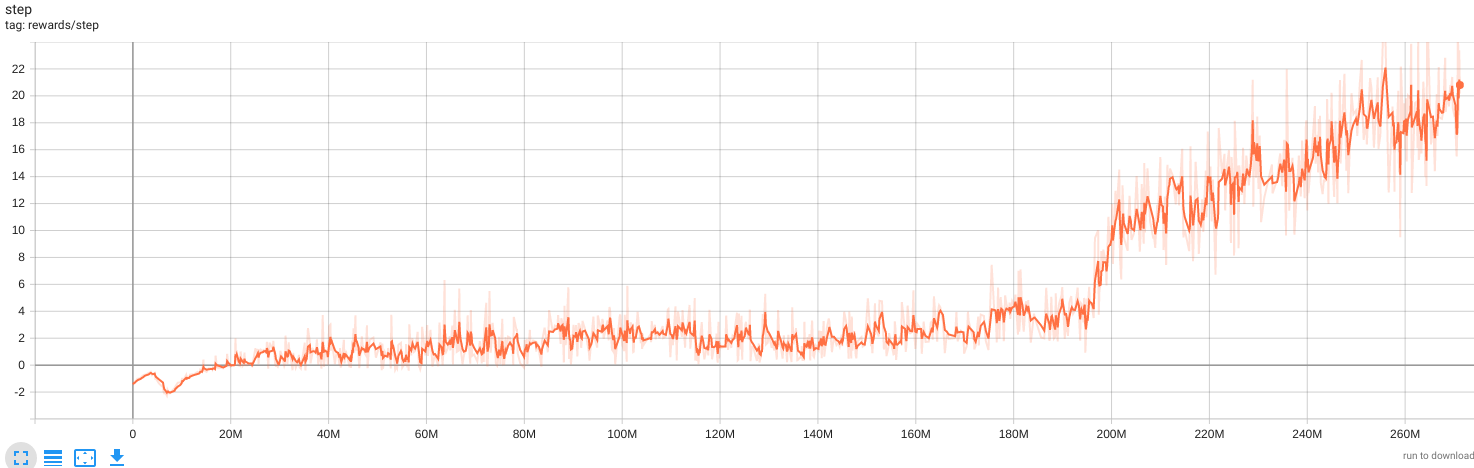}
            \caption{Kicking Motion Training Graph }
            \label{fig:kick_train}
            \end{figure}

        After Bez successfully learned how to kick the ball without falling, their progress appeared to plateau, as indicated by the figure \ref{fig:kick_train} before 180 million steps. Although Bez had acquired the skill to kick a ball, they lacked accuracy.
        \\\\\
        To address this issue and encourage the policy to prioritize accuracy over ball velocity, a curriculum training approach was implemented. The approach involved building upon previously trained policies with additional punishments to ensure the ball's direction fell within a specific range of the target location. Additional boundary conditions were introduced to limit its forward movement after kicking the ball. However, some of these conditions became too restrictive over time, leading to weaker kicks. To address this, the conditions were relaxed, and Bez was retrained with a new set of boundary conditions, resulting in stronger kicks.
        \\\\\
        By using this approach, inaccurate and unstable kicks within the search space were significantly reduced, leading to a significant increase in the number of powerful and accurate kicks that hit their intended target. As a result, Bez received more rewards for accurate kicks indicated by figure \ref{fig:kick_train} after 180 million steps.
        
    \subsection{Walk}
        \subsubsection{Description}
            The walking task involves directing Bez to walk toward a specified goal location, which is set at the start of each episode. To successfully complete the task, Bez must navigate from its initial position to the target location with precision, without deviating from the designated path or losing balance.
        \subsubsection{Reward Engineering}
            The task reward at a high-level incentives Bez to move forward as quickly as possible by assigning the $VelocityForwardReward$, which is based on the relative distance between Bez and the goal. The reward formula is as follows:
        
            \[DistanceToGoal = GoalPosition - BezPosition\]
            \[VelocityForwardReward =\frac{DistanceToGoal}{\|DistanceToGoal \|} \cdot IMULinearVelocityXY \]
        
            This reward function encourages Bez to run at maximum speed towards the goal, maximizing their forward velocity. The $IMULinearVelocityXY$ term represents the velocity of Bez in the x-y plane, and the reward is calculated by taking the dot product of this velocity with the normalized vector pointing from Bez to the goal. The resulting scalar value represents the forward component of Bez's velocity, and higher values indicate a faster movement toward the goal.
        \subsubsection{Curriculum}
            Once Bez had mastered walking in a straight line without falling, the next step was to add a new challenge to the training. The original fixed goal was replaced with a randomized goal, situated within a 2 x 2-meter box around Bez's initial location. This modification allowed the walking controller to become more versatile and adaptable to different types of movements.
            \\\\\
            By introducing a randomized goal, Bez's walking ability was tested in a range of scenarios, which helped the walking controller become more generalized. This increased adaptability is beneficial in real-world applications where environmental conditions are unpredictable and constantly changing.
            \\\\\
            Overall, the addition of a randomized goal was a valuable enhancement to the training process, enabling Bez's walking controller to be more effective and efficient in a variety of contexts. 
    \subsection{Jump}
        \subsubsection{Expert Demonstration}
            \begin{figure}[h]
            \centering
            \includegraphics[width=75mm,scale=0.5]{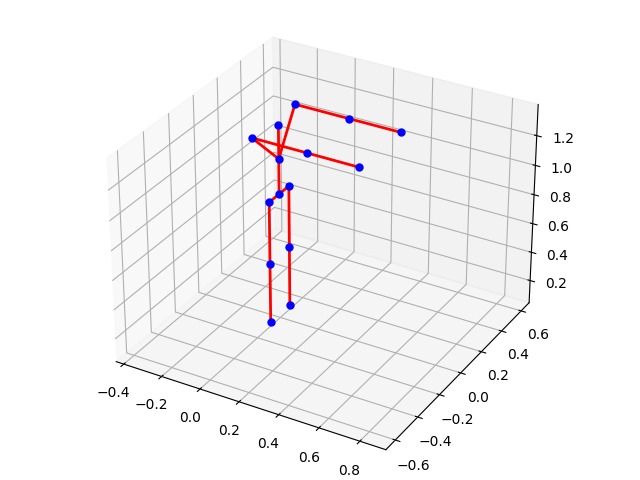}
            \includegraphics[width=75mm,scale=0.5]{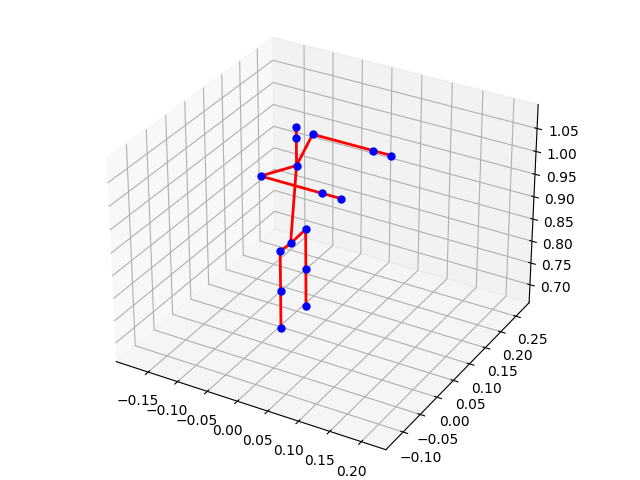}
            \caption{Humanoid Skeleton structure (left) and Bez skeleton structure (right) in T-Pose}
            \label{fig:tpose}
            \end{figure}
            \begin{figure}[h]
            \centering
            \includegraphics[width=75mm,scale=0.5]{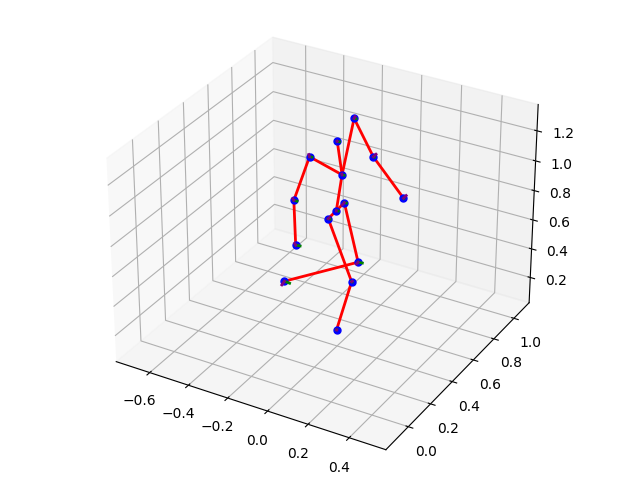}
            \includegraphics[width=75mm,scale=0.5]{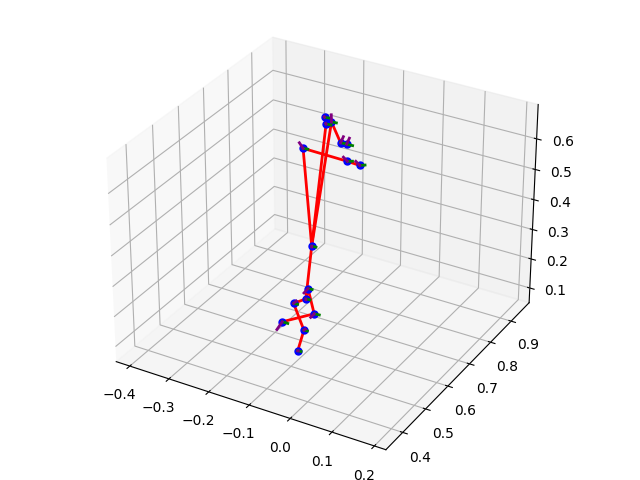}
            \caption{Humanoid Skeleton structure (left) and Bez skeleton structure (right) while walking}
            \label{fig:walk}
            \end{figure}
            Since the other skills use reward engineering they can produce infeasible motions. A way to circumvent this drawback is to use imitation learning or AMP to train the policy a base motion and improve towards optimization. The Motion capture data of humans walking, running, and performing various tasks were collected from the Carnegie Mellon University (CMU) and Simon Fraser University (SFU) databases due to their vast and diverse collection \cite{cmu}, \cite{sfu}. However, a challenge in using these datasets was that they were captured from human subjects with a skeleton structure of 52 joints, while Bez has only 18 joints. To overcome this, the motion capture data were retargeted using Poselib to Bez's skeleton structure. This allowed for the transfer of human motion to Bez, resulting in more natural and human-like movements \cite{makoviychuk2021isaac}. An example of the retargeted motion can be found below in figure \ref{fig:tpose} and \ref{fig:walk}.
        \subsubsection{Rewards}
            The reward function is limited to only imitating the motion references. There is no task-specific reward function

\newpage
\section{Skill Analysis}
    \subsection{Experiments}
     Experiments were conducted to evaluate the impact of foot pressure sensors on the performance and stability of Bez's kicking behavior. Two different versions of Bez's URDF were used: one with foot pressure sensors and one without. The results showed that the foot pressure sensors had a negative effect on Bez's performance and stability, as their design weakened the foot and caused Bez to fall more frequently over time.
     \\\\\
     To optimize Bez's performance, various parameters of the reward and reset functions were adjusted, including the height threshold for determining if Bez had fallen, the scaling of different components of the reward function to prioritize velocity or stability, the sensitivity of the foot pressure sensors, and the magnitude of the rewards for failed and successful episodes. Similar experiments were conducted for the walking task.
     \\\\\
     To see a demo of Bez tasks click on the hyperlink for 
     \href{https://drive.google.com/file/d/1Cyms_T3oWZoMaW69t-5PMxBNRFpFf_eY/view?usp=sharing}{Kicking},
     \href{https://drive.google.com/file/d/19GCgfvvGYjweLEn7ZBMhO4SBAYYqVL6E/view?usp=sharing}{Walking}, and \href{https://drive.google.com/file/d/1kOQWXYKEsBclPH4ebD-uuuZmzzrWTaHk/view?usp=sharing}{Jumping}

    \subsection{Evaluation Metrics}
        \subsubsection{Common Metrics}
            To evaluate a policy's performance, metrics emphasizing stability are commonly used. Pitch Angle measures Bez's lean by evaluating the angle between the ground plane and Bez's head orientation around the Y-axis, indicating the motion's stability. Fell (Out of 10) measures the percentage of falls that occurred during the motion out of 10 attempts, indicating the policy's effectiveness in avoiding risky movements.
        \subsubsection{Kick Metrics}
            For the kicking motion, specific metrics are Kick velocity and Accuracy (Out of 10). Kick speed measures the strength and power of the kick, while Accuracy (Out of 10) evaluates the precision and consistency of the motion by measuring the percentage of successful attempts that reach the goal area of (1,0). These metrics provide important insights for improving the effectiveness of the kicking motion.
        \subsubsection{Walk Metrics}
            To evaluate the walking motion, Average Velocity for straight and randomized goals are used. It measures the average speed of Bez's movement toward both goals. This metric provides insights into the overall speed of the walking motion and can be used to identify areas that may require improvement.
        \subsubsection{Jump Metrics}
            When evaluating the jumping motion, several metrics are typically used. These include Distance X and Distance Z, which measure the horizontal and vertical distances traveled by Bez during the jump, respectively. Velocity X and Velocity Z measure the speed of Bez's movement in the horizontal and vertical directions, respectively.
            \\\\\
            Additionally, Average Jumps in a Row is a metric that indicates the number of consecutive jumps that Bez can perform. This metric provides insight into the stability of the jumping motion and can be used to identify areas for improvement.
    \newpage
    \subsection{Evaluation Results}
        One of the main objectives of this thesis is to demonstrate that RL policies outperform previous methods in various soccer skills. The section presents quantitative analyses to support the evidence and create a pipeline for controllers for a wide variety of skills.

        \begin{table}[h!]
        \caption{Evaluation of Kick motion policy}
        \centering
        \begin{tabular} {||c | c | c||} 
         \hline
        Metric  & Trained Policy & Previous Method  \\
         \hline
         Kick Velocity & 1.3 m/s & 0.3 m/s  \\
         \hline
         Accuracy (Out of 10)  & 90\% & 40\% \\
         \hline
         Pitch Angle  & 7$^{\circ}$ & 3$^{\circ}$ \\
        \hline
        Fell (Out of 10)  & 20\% & 0\%  \\
        \hline
        
        \end{tabular}   
        \label{table:kick}
        \end{table}

        \begin{table}[h!]
        \caption{Evaluation of Walking motion policy}
        \centering
        \begin{tabular} {||c | c | c| c||} 
         \hline
        Metric  & Trained Policy & Previous Method & LORM \cite{lorm} \\
         \hline
         Avg Velocity (Straight) & 1.0 m/s & 0.1 m/s & 0.488 m/s\\
         \hline
         Avg Velocity (Random) & 0.55 m/s & 0.1 m/s & 0.488 m/s\\
         \hline
         Pitch Angle  & 6$^{\circ}$ & 3$^{\circ}$ & N/A\\
        \hline
        Fell (Out of 10)  & 20\% & 0\%  & 0\%\\
        \hline
        
        \end{tabular}   
        \label{table:walk}
        \end{table}\
        \\\\\
        Tables \ref{table:kick} and \ref{table:walk} demonstrate that the RL policies developed for the kicking and walking motions exhibit a significant improvement in performance compared to previous methods used on Bez, especially in terms of speed and accuracy. It is worth noting that there is a slight compromise in stability as indicated by a higher pitch angle and fall percentage, but the performance gains more than compensate for them. By comparing the results of LORM with the policies developed in this thesis, Table \ref{table:walk} confirms that both policies show comparable performance, but the thesis policy outperforms LORM's trained policy by a reasonable margin. Comparable results for kicking and jumping could not be found.
        \\\\\
         \begin{table}[h!]
        \caption{Evaluation of Jump motion policy}
        \centering
        \begin{tabular} {||c | c ||} 
         \hline
        Metric  & Trained Policy   \\
         \hline
         Distance X & 0.8 m  \\
         \hline
         Distance Z  & 0.35 m  \\
         \hline
         Velocity X & 1.82 m/s  \\
         \hline
         Velocity Z  & 1.51 m/s  \\
         \hline
         Avg Jumps in a Row & 6  \\
         \hline
         Pitch Angle  & 6.5$^{\circ}$  \\
        \hline
        Fell (Out of 10)  & 10\%  \\
        \hline
        
        \end{tabular}   
        \label{table:jump}
        \end{table}
         Table \ref{table:jump} shows the performance of the jumping motion, which has the best stability among all the models, highlighted by its low fall percentage and the number of jumps successfully done in a row. This further emphasizes the power of AMP.
        \\\\\
        In addition to creating RL policies, another goal is to create a pipeline for controllers with a wide variety of skills. Table \ref{table:train} presents the memory usage and training time for each model based on the computer specs mentioned. These results serve as evidence that policies of different complexity can be created with a consumer-grade computer and very fast training times.
         \begin{table}[h!]
        \caption{Training Specification}
        \centering
        \begin{tabular} {||c | c | c||} 
         \hline
        Motion type  & Time to train & Ram usage (4000 Env) \\
         \hline
         Kicking & 49 minutes & 8GB\\
         \hline
         Walking & 74 minutes & 8GB  \\
         \hline
         Jumping & 44 minutes & 8GB\\
        \hline
        
        \end{tabular}   
        \label{table:train}
        \end{table}
        \\\\\
        It is essential to mention the limitations of this thesis, as it was conducted in a simulation environment. Real-world scenarios could differ, affecting the policies' effectiveness.
    \newpage
    \subsection{Analysis of Curriculum Training}
        One of the main objectives of this thesis is to develop RL policies that are more dynamic and adaptive to the environment. To achieve this, a curriculum training approach was employed, which enables us to leverage simpler learned skills as a foundation to tackle more complex and challenging tasks. Doing so can make the policies more accurate, generalized, and adaptive to different environmental conditions.
        \\\\\
        For instance, after successfully training the robot to perform a simple kick and walk forward, leveraged these skills as a foundation to train the robot to kick to a specific location within a 1-meter distance with a 5$^{\circ}$ angle variance. Moreover, the robot was trained to walk in any direction within a 2x2 box, which significantly improved the robot's adaptability and generalization skills.
        \\\\\
        As a result of the curriculum training approach, observed significant improvements in the accuracy and generalization skills of the robot controllers. For example, the accuracy of the kicking policy increased from 20\% to 90\%, even though the kick velocity decreased from 1.5m/s to 1.3m/s. Similarly, the robot's walking velocity was halved, but the ability to walk in any direction made it more usable and adaptable to different environmental conditions.
    \subsection{Feasible motions}
        The kicking and walking motions in this project were designed using a reward engineering approach, where the desired behaviors for the motion were defined through careful experimentation and tuning of the equations and weights within the reward function. This process required 33 experiments for the kicking motion and 6 for the walking motion. However, a significant downside of reward engineering is that it can result in complex rewards that lead to unnatural and infeasible motions.
        \\\\\
        \begin{figure}%
            \centering
            \subfloat[\centering]{{\includegraphics[width=2.5cm]{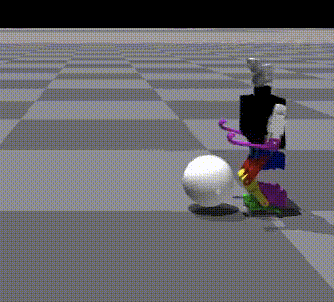} }}%
            \subfloat[\centering ]{{\includegraphics[width=2.5cm]{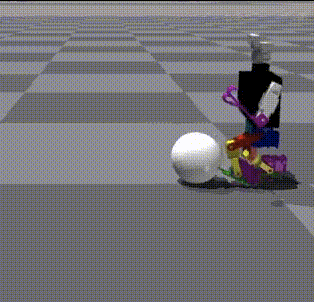} }}%
            \subfloat[\centering ]{{\includegraphics[width=2.5cm]{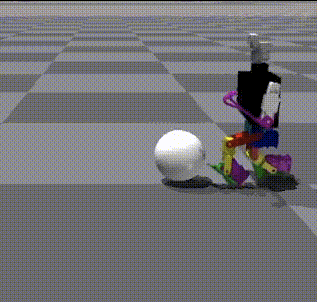} }}%
            \subfloat[\centering ]{{\includegraphics[width=2.5cm]{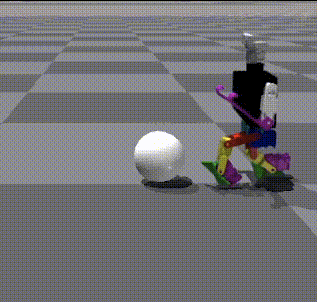} }}%
            \subfloat[\centering ]{{\includegraphics[width=2.5cm]{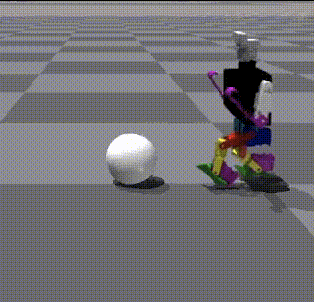} }}%
            \subfloat[\centering]{{\includegraphics[width=2.5cm]{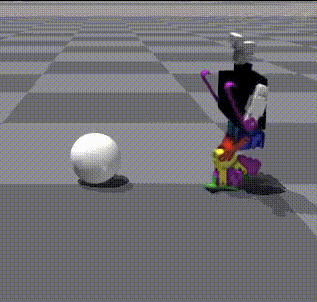} }}%
            \caption{Sequence of the robot during the kick}%
            \label{fig:kicking}%
        \end{figure}
        \begin{figure}%
            \centering
            \subfloat[\centering]{{\includegraphics[width=2.5cm]{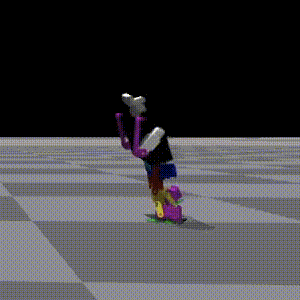} }}%
            \subfloat[\centering ]{{\includegraphics[width=2.5cm]{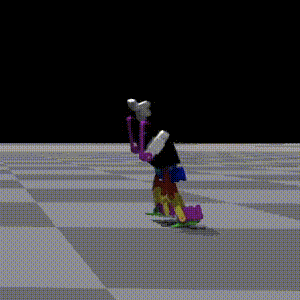} }}%
            \subfloat[\centering ]{{\includegraphics[width=2.5cm]{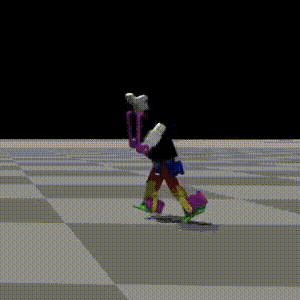} }}%
            \subfloat[\centering ]{{\includegraphics[width=2.5cm]{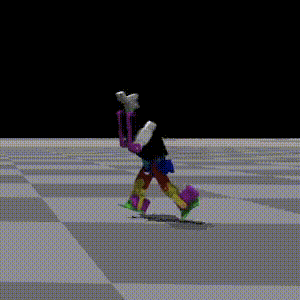} }}%
            \subfloat[\centering ]{{\includegraphics[width=2.5cm]{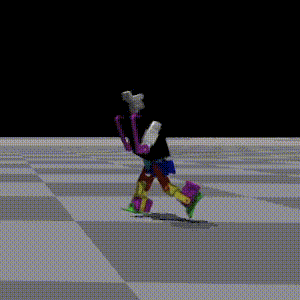} }}%
            \subfloat[\centering]{{\includegraphics[width=2.5cm]{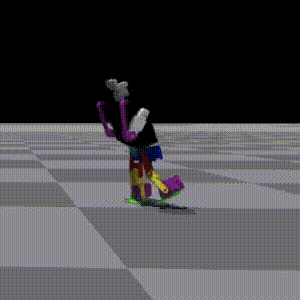} }}%
            \caption{Sequence of the robot during the walk}%
            \label{fig:walking}%
        \end{figure}
        \begin{figure}%
            \centering
            \subfloat[\centering]{{\includegraphics[width=2.5cm]{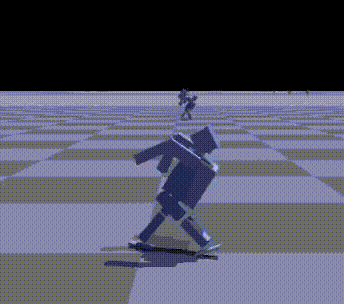} }}%
            \subfloat[\centering ]{{\includegraphics[width=2.5cm]{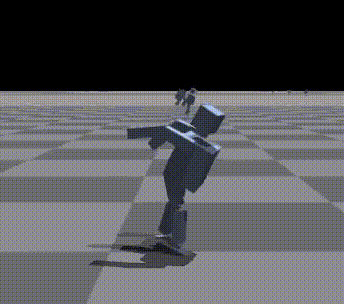} }}%
            \subfloat[\centering ]{{\includegraphics[width=2.5cm]{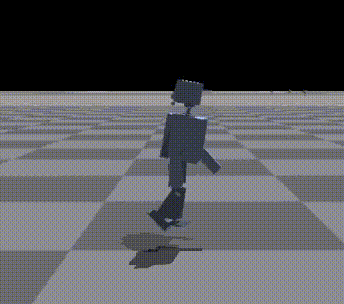} }}%
            \subfloat[\centering ]{{\includegraphics[width=2.5cm]{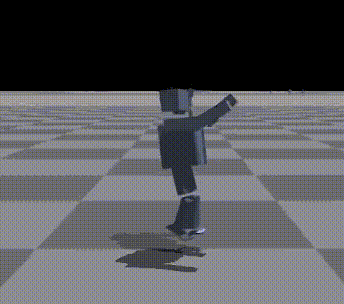} }}%
            \subfloat[\centering ]{{\includegraphics[width=2.5cm]{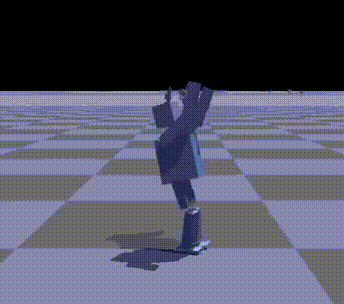} }}%
            \subfloat[\centering]{{\includegraphics[width=2.5cm]{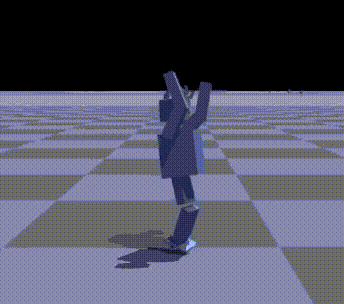} }}%
            \caption{Sequence of the robot during the jump}%
            \label{fig:jumping}%
        \end{figure}
        
        As shown in the frame-by-frame figures \ref{fig:kicking} \ref{fig:walking} \ref{fig:jumping}, the kicking and walking motions are not always feasible, and they may not transfer well to different environments without significant tuning and retraining. These limitations highlight the need for more dynamic and adaptive RL policies that can learn from the environment without requiring extensive manual tuning.
        \\\\\
        AMP is a promising approach that offers solutions to the limitations encountered in traditional manual engineering of reward functions. By utilizing AMP, policies can learn more natural and feasible motions, which result in greater adaptability to diverse environmental conditions. The use of AMP also allows for greater flexibility in the agent's movements, as exemplified by the jumping motion. Unlike constrained and unnatural kicking and walking motions, the jumping motion can be easily adapted to different scenarios and environments. Thus, the use of AMP in training policies offers a more effective and efficient way to optimize agent behavior in dynamic and uncertain environments.

    \subsection{Sim2Real}
        Domain randomization is a widely-used technique in training generalized policies that can adapt to diverse environments. However, during my thesis project, I encountered significant difficulties in transferring the learned policies to new environments due to significant disparities between the training and testing domains.

\newpage
\section{Conclusion}
    In this thesis, a more efficient and effective approach was developed for training control-related tasks such as kicking, walking, and jumping for humanoid robots, which overcomes the limitations of traditional Reinforcement Learning methods and adapts to real-world environments. The use of curriculum training resulted in the development of RL policies that efficiently learned various skills and outperformed previous methods. Additionally, the use of Adversarial Motion Priors (AMP) technique offered greater flexibility and freedom in developing new behaviors for achieving task objectives while requiring less complex reward functions. This created a more reliable model and reduced the likelihood of producing undesirable movements.
    \\\\\
    Furthermore, the developed RL policies were found to be more dynamic and adaptive to the environment, which has significant implications for the development of more sophisticated and intelligent robotic systems that can perform complex tasks in real-world environments. However, the transfer of a learned policy from simulation to the real world was unsuccessful, highlighting the limitations of current RL methods in fully adapting to real-world scenarios.
    \\\\\
    Overall, this research has contributed to advancing the field of robotics by offering a more efficient and effective approach for training control-related tasks for humanoid robots. Demonstrating the effectiveness of curriculum training and AMP in developing more efficient and adaptive RL policies for humanoid robots. While there are limitations to the current methods, this research provides a foundation for future work in developing more sophisticated RL policies that can better adapt to real-world environments.

\newpage
\section{Future Work}
    Future work in this field should focus on improving the transferability of the learned policies from simulation to the real-world environment. One approach that could be explored is the use of domain randomization or a low-pass filter for the actions, as well as domain adaptation techniques, to bridge the gap between the two environments. Hyperparameter optimization could also be performed for the best-performing state-action combinations to further improve the learning process.
    \\\\\
    An ablation study could also be conducted to determine which factors are most important in the observation state vector. This could help to identify the critical features of the environment that the RL algorithm should focus on for better performance.
    \\\\\
    In addition, Long Short-Term Memory (LSTM) networks could be used to achieve better and faster response times with predictive capabilities. This would enable the robot to anticipate future states and actions more accurately and react more quickly to changes in the environment.
    \\\\\
    More work could also be done with the Adversarial Motion Priors (AMP) technique to develop more efficient and adaptable motions for the robot. Future research could explore how to maintain a consistent style while achieving more complex motions. This technique could also be extended to other skills such as rotation, dribble kicking, multi-direction kicking, and goalie defense.
    \\\\\
    Overall, there is still much work to be done in developing more sophisticated and efficient RL policies for humanoid robots. Further research and testing are needed to evaluate the effectiveness of these approaches in real-world scenarios and to identify the most critical factors for achieving success in complex control-related tasks.
 \newpage   
\bibliographystyle{ieeetr}
\bibliography{sample}
\newpage
\vspace*{5cm}

\section*{\begin{center}Appendix\end{center} }
\appendix
\newpage
\section{Codebase}
Link to \href{https://github.com/utra-robosoccer/Bez_IsaacGym}{github}, 
\newpage
\section{Ready Joint Angles}
   \begin{table}[h!]
        \caption{Ready Joint Angles}
        \centering
        \begin{tabular} {||c | c ||} 
         \hline
        Joint Name  & Joint Value (rad)   \\
         \hline
        right\_leg\_motor\_0 & 0.0 \\
         \hline
    right\_leg\_motor\_1 & 0.0 \\
         \hline
    right\_leg\_motor\_2 & 0.564 \\
         \hline
    right\_leg\_motor\_3 & -1.176 \\
         \hline
    right\_leg\_motor\_4 & 0.613 \\
         \hline
    right\_leg\_motor\_5 & 0.0 \\
         \hline
    left\_leg\_motor\_0 & 0.0 \\
         \hline
    left\_leg\_motor\_1 & 0.0 \\
         \hline
    left\_leg\_motor\_2 & 0.564 \\
         \hline
    left\_leg\_motor\_3 & -1.176 \\
         \hline
    left\_leg\_motor\_4 & 0.613 \\
         \hline
    left\_leg\_motor\_5 & 0.0 \\
         \hline
    right\_arm\_motor\_0 & 0.0 \\
         \hline
    right\_arm\_motor\_1 & 1.5 \\
         \hline
    left\_arm\_motor\_0 & 0.0 \\
         \hline
    left\_arm\_motor\_1 & 1.5 \\
         \hline
    head\_motor\_0 & 0.0 \\
         \hline
    head\_motor\_1 & 0.0 \\
         \hline
        \end{tabular}   
        \label{table:jump}
        \end{table}
\newpage

\section{Kick Experiments}
   \begin{table}[h!]
        \caption{Kicking Motion Experiments}
        \centering
         \begin{adjustbox}{width=\textwidth}
        \begin{tabular} {||p{0.05\linewidth} | p{0.35\linewidth} | p{0.6\linewidth}||}   
         \hline
          \#  & Changes & Notes   \\
         \hline
         1 & Setting up & \\
         \hline
         2  & Setting up & \\
         \hline
         3 & Setting up  & \\
         \hline
         4  & Setting up  & \\
         \hline
         5 & no ground reward & \\
         \hline
         6  & 0.01 \* ground &  \\
        \hline
        7  & new feet and 0.01 \* ball velocity  &  \\
        \hline
        8  & 5 but optimized &  \\
        \hline
        9  & 6 but optimized  &  \\
        \hline
        10  & 8 + new feet &  \\
        \hline
        11  & 9 + new feet &  \\
        \hline
        12  & 8 + revert ball scale &  walk forward too much maybe limit position movement. Reverting ball scale seems more balanced but slower kick\\
        \hline
        13  & 12 + linear velocity after 0.1m stop moving & Does not work. stops movement but worsens stabilization \\
        \hline
        14  &  12 + new velocity calculations & Same but a little worse \\
        \hline
        15  & 13 + feet & Stabilization is a little worse but does attempt flat feet more \\
        \hline
        16  & 13 + normalize feet scale to 0.01  & I don't think norm works \\
        \hline
        17  & 15 + scale feet 0.01 =\> 0.02  & more stable \\
        \hline
        18  &  15 + old feet & cleats does not work \\
        \hline
        19  & 13 + fall punishment 1 =\> 2.5  & results in faster kicks \\
        \hline
        20  & 13 + new bounding functions for ball angle = 0.3  & maybe not big enough \\
        \hline
        21  & 13 + new bound function for ball angle using vectors  & much better \\
        \hline
        22  & 21 + new bound function for ball angle based on distance + checkpoint & better \\
        \hline
        23  & 22 + bound = 0.3 & dof velocity does not work its too restricting \\
        \hline
        24  & 23 + win =\> 100 -(100 -time) &  better \\
        \hline
        25  & 24 + feet &  \\
        \hline
        26  & 24 + new dist height + 10* death scale & too restricting \\
        \hline
        27  & 25 + bound = 1.57 &  better but less accurate\\
        \hline
        28  & 24 + variable bound &  \\
        \hline
        29  & 28 + bound = 0.01 &  \\
        \hline
        30  & 29 + bound = 0.008 &  \\
        \hline
        31  & 30 + bound = 0.009 &  \\
        \hline
        32  & 24 + var bound & Best option \\
        \hline
        33  & 25 + var bound &  \\
        \hline
        
        \end{tabular} 
        \end{adjustbox}
        \label{table:exp}
        \end{table}
\section{Walk Experiments}
   \begin{table}[h!]
        \caption{Walking Motion Experiments}
        \centering
         \begin{adjustbox}{width=\textwidth}
        \begin{tabular} {||p{0.05\linewidth} | p{0.35\linewidth} | p{0.6\linewidth}||}   
         \hline
          \#  & Changes & Notes   \\
         \hline
         1 & 0.1 velocity forward & \\
         \hline
         2  & 0.5 velocity forward & \\
         \hline
         3 & 1 velocity forward and change win state bounds 0.01 =\>0.1 & best balance\\
         \hline
         4  & 10 velocity forward  & Most like a sprint\\
         \hline
         5 & 3 + 100*goal + new distance height & \\
         \hline
         6  & 4 + 100*goal + new distance height &  \\
        \hline

        \end{tabular} 
        \end{adjustbox}
        \label{table:exp}
        \end{table}
\newpage
\section{Training graph}
    \begin{figure}[h]
                \centering
                \includegraphics[width=150mm,scale=1]{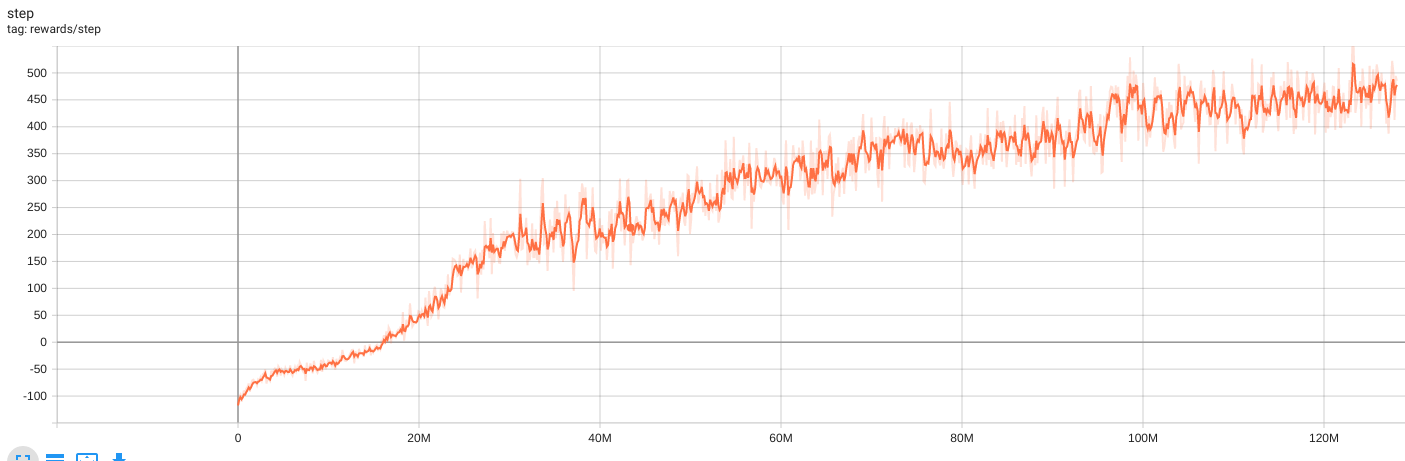}
                \caption{Walking Motion Training Graph Random Goal }
                \label{fig:sim}
                \end{figure}
    
    \begin{figure}[h]
                \centering
                \includegraphics[width=150mm,scale=1]{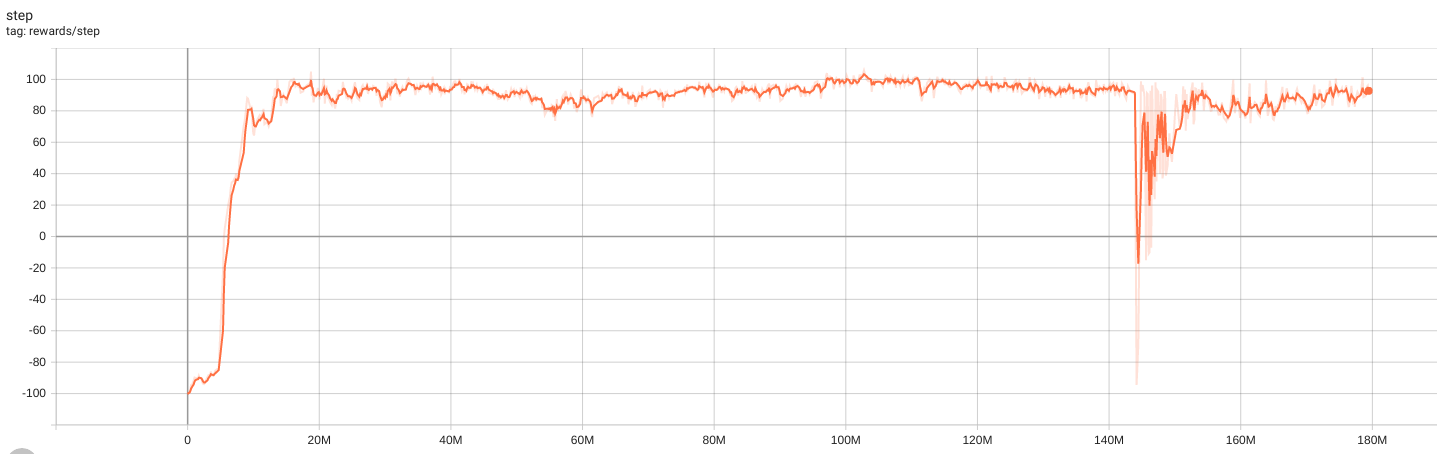}
                \caption{Walking Motion Training Graph Straight Goal }
                \label{fig:sim}
                \end{figure}
    
    \begin{figure}[h]
                \centering
                \includegraphics[width=150mm,scale=1]{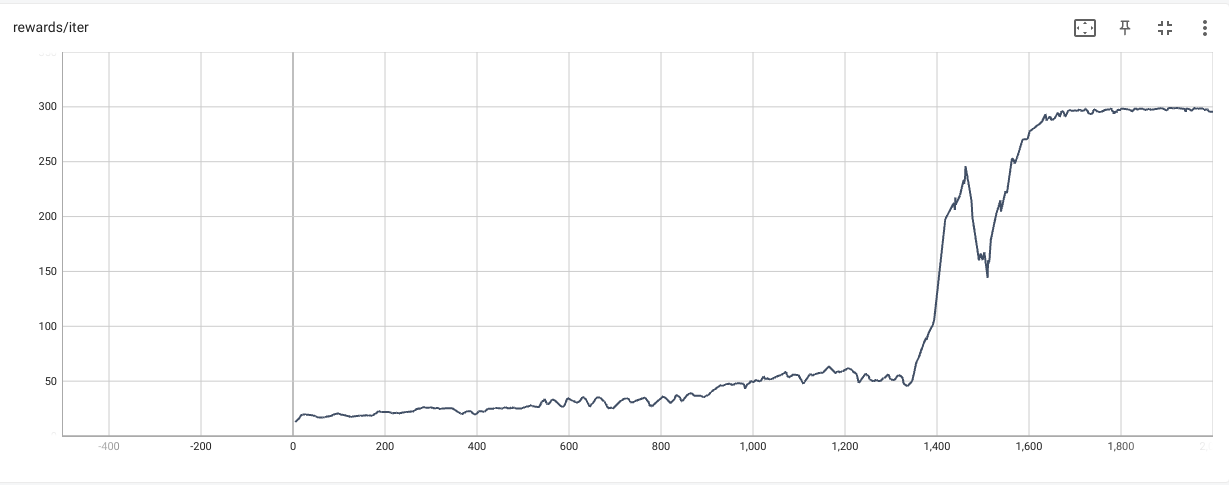}
                \caption{Jumping Motion Training Graph }
                \label{fig:sim}
                \end{figure}
\afterpage{\null\newpage}

\end{document}